\documentclass[10pt,twocolumn,letterpaper]{article}

\usepackage{cvpr}
\usepackage{pifont}
\usepackage{times}
\usepackage{epsfig}
\usepackage{graphicx}
\usepackage{amsmath}
\usepackage{amssymb}
\usepackage{subcaption}
\usepackage{booktabs}
\usepackage{threeparttable}
\usepackage{textcomp}

\usepackage[breaklinks=true,bookmarks=false]{hyperref}

\cvprfinalcopy 

\def\cvprPaperID{****} 

\begin{document}

\title{The DeepFake Detection Challenge (DFDC) Dataset}
\author{Brian Dolhansky, Joanna Bitton, Ben Pflaum, Jikuo Lu,\\
Russ Howes, Menglin Wang, Cristian Canton Ferrer\\
\\
Facebook AI\\
\vspace{-1cm}
}


\makeatletter
\let\@oldmaketitle\@maketitle
\renewcommand{\@maketitle}{\@oldmaketitle
  \includegraphics[width=\linewidth]
    {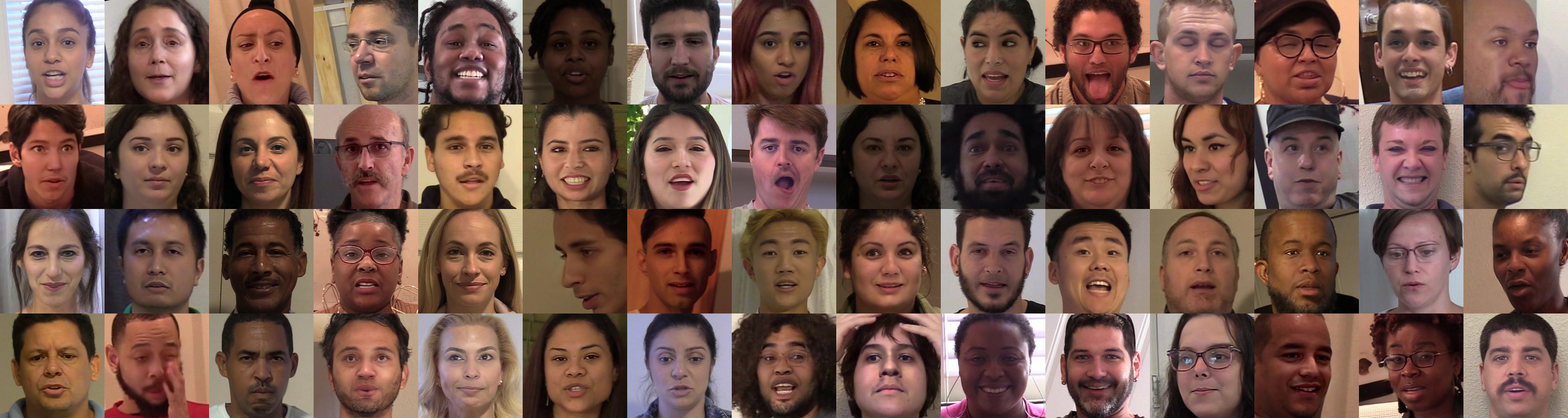}\bigskip\bigskip}
\makeatother

\newcommand{\cmark}{\ding{51}}%
\newcommand{\xmark}{\ding{53}}%

\maketitle
\begin{abstract}
Deepfakes are a recent off-the-shelf manipulation technique that allows anyone to swap two identities in a single video. In addition to Deepfakes, a variety of GAN-based face swapping methods have also been published with accompanying code. To counter this emerging threat, we have constructed an extremely large face swap video dataset to enable the training of detection models, and organized the accompanying \textit{DeepFake Detection Challenge} (DFDC) Kaggle competition. Importantly, all recorded subjects agreed to participate in and have their likenesses modified during the construction of the face-swapped dataset. 

The DFDC dataset is by far the largest currently- and publicly-available face swap video dataset, with over 100,000 total clips sourced from 3,426 paid actors, produced with several Deepfake, GAN-based, and non-learned methods. In addition to describing the methods used to construct the dataset, we provide a detailed analysis of the top submissions from the Kaggle contest. We show although Deepfake detection is extremely difficult and still an unsolved problem, a Deepfake detection model trained only on the DFDC can generalize to real "in-the-wild" Deepfake videos, and such a model can be a valuable analysis tool when analyzing potentially Deepfaked videos. Training, validation and testing corpuses can be downloaded from \url{https://ai.facebook.com/datasets/dfdc}.
\end{abstract}

\section{Introduction}
Swapping faces in photographs has a long history, spanning over one hundred and fifty years~\cite{farid2011photo}, as film and digital imagery have a powerful effect on both individuals and societal discourse~\cite{jolly2003fake}. Previously, creating fake but convincing images or video tampering required specialized knowledge or expensive computing resources~\cite{sturman1994brief}. More recently, a new technology called \textit{Deepfakes}\footnote{The term "Deepfake" has multiple definitions, but we define a Deepfake as a video containing a swapped face \textit{and} produced with a deep neural network. This is constrasted with so-called "cheapfakes" - if a fake video was produced with machine learning, it is a Deepfake, whereas if it was created with widely-available software with no learning component, it is a cheapfake~\cite{paris2019deepfakes}.}
has emerged~\cite{verdoliva2020media} - a technology that can produce extremely convincing face-swapped videos. Producing a Deepfake does not require specialized hardware beyond a consumer-grade GPU, and several off-the-shelf software packages for creating Deepfakes have been released. The combination of these factors has lead to an explosion in their popularity, both in terms of producing parody videos for entertainment, and for use in targeted attacks against individuals or institutions~\cite{floridi2018artificial}.

With the understanding that it is now possible for a member of the general public to automatically create convincing fake face-swapped videos with simple hardware, the need for creating automated detection methods becomes clear~\cite{chesney2019deep}. While digital forensics experts can analyze single, high-impact videos for evidence of manipulation, this cannot scale to reviewing each of the hundreds of thousands of videos uploaded to Internet or social media platforms every day. Detecting Deepfakes at scale necessitates scalable methods, and computer vision or multimodal models are particularly suited to this challenge. However, these models require training data, and even though it is possible to create several convincing Deepfakes easily, the cost of producing the hundreds of thousands of Deepfake videos necessary to train these models is often cost prohibitive. In order to accelerate advancements in the state of the art of Deepfake detection, we have constructed and publicly released the largest Deepfake detection dataset to date.

Our first major contribution is the DeepFake Detection Challenge (DFDC) Dataset. Motivated primarily by the fact that many previously-released datasets contained few videos with few subjects and with a limited size and number of methods represented, we wanted to release a dataset with a large number of clips, of varying quality, and with a good representation of current state of the art face swap methods. Furthermore, as we observed that many publicly-released datasets~\cite{Jiang2020, Korshunov2018, Li2019, roessler2019faceforensicspp, Yang2019} did not guarantee that their subjects were willing participants or agreed to have their faces modified, we solicited video data from 3,426 paid actors and actresses speaking in a variety of settings for roughly 15 minutes each. All participants agreed to appear in a dataset where their faces may be manipulated by a computer vision algorithm. The DFDC Dataset is both the largest currently-available Deepfake dataset, and one of only a handful of datasets containing footage recorded specifically for use in machine learning tasks (the others being the much smaller Google Deepfake Detection Dataset~\cite{dufour_gully_2019} and the preview version of this dataset~\cite{Dolhansky2019}).

Beyond building and releasing a dataset, our second major contribution is a now-completed benchmark competition using this data\footnote{https://www.kaggle.com/c/deepfake-detection-challenge}, and the resulting analysis. The benefits of a competition of this size are many. First, the monetary prizes provided a large incentive for experts in computer vision or Deepfake detection to dedicate time and computational resources to train models for benchmarking. Second, hosting a public competition obviates the need for the authors of a paper to train and test a model on a dataset they produced. Releasing a dataset and a benchmark simultaneously can introduce bias, as the creators of a dataset have intimate knowledge of what methods were used while constructing the dataset. Third, gathering thousands of submissions and running them against real Deepfake videos \textit{that participants never see} paints an extremely accurate picture of the true Deepfake detection state of the art. 

\section{Previous work}

\begin{figure}
    \centering
    \includegraphics[width=\linewidth]{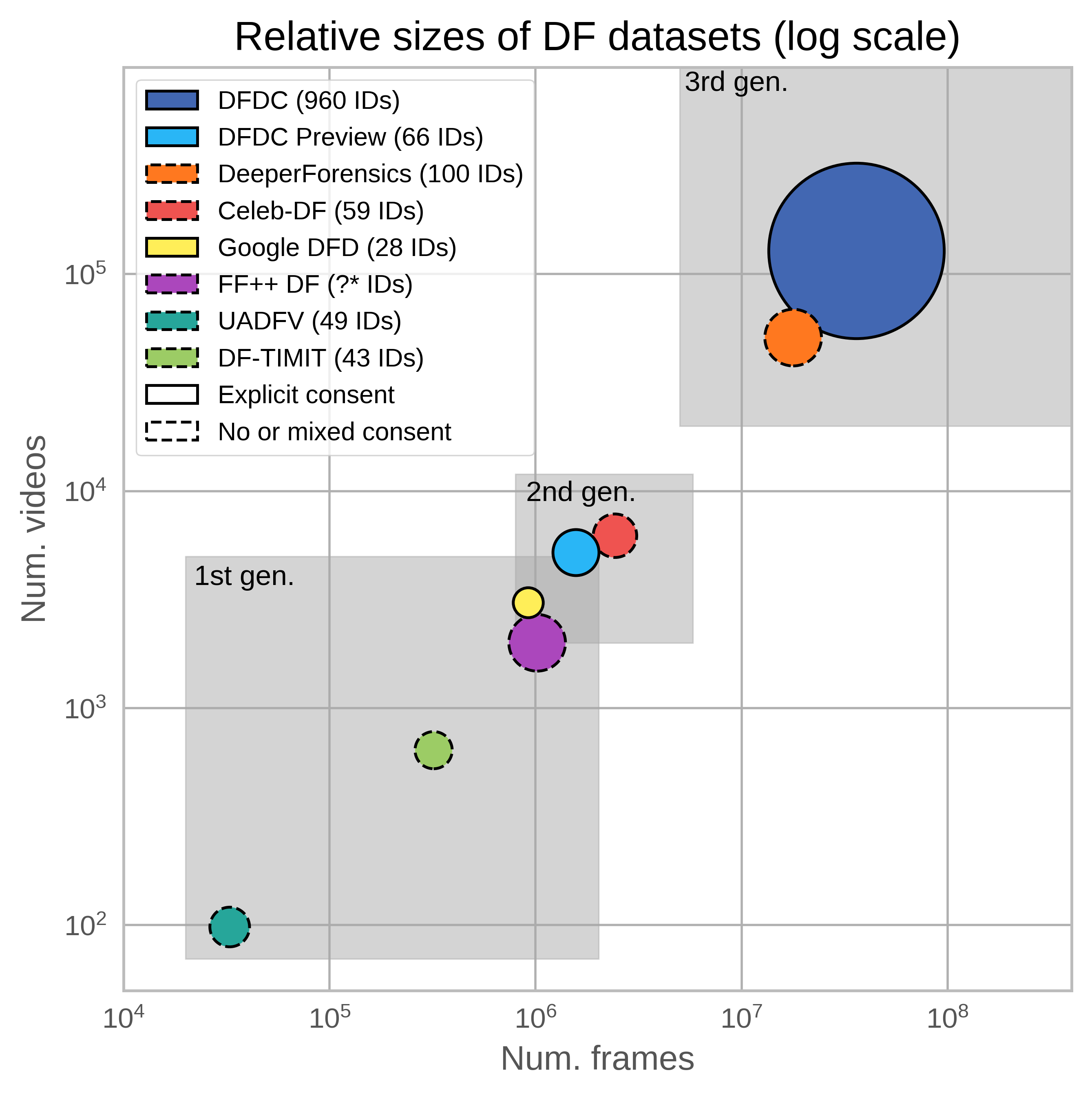}
    \caption{Comparison of current Deepfake datasets. Both axes are shown in log scale - the DFDC is over an order of magnitude larger than any other available dataset, both in terms of the number of frames and number of videos. Rough boundaries for dataset "generation" (as given in~\cite{Li2019}) also shown. Overlapping circles do not indicate inclusion; circle sizes are merely a visualization of the number of fake identities present in the dataset.}
    \label{fig:datasets}
\end{figure}

\begin{table*}[t]
  \small
  \centering
  \begin{threeparttable}
    \caption{Quantitative comparison of various Deepfake datasets}
      \begin{tabular}{l|rrcrrrrr}
      \hline
      Dataset & \shortstack[r]{Unique\\fake videos} & \shortstack[r]{Total \\ videos} & \shortstack[r]{Unclear\\rights} & \shortstack[r]{Agreeing\\subjects\tnote{a}} & \shortstack[r]{Total\\subjects} & Methods & No. perturb. & \shortstack[r]{No.\\benchmarks\tnote{b}}\\
      \hline
        DF-TIMIT~\cite{Korshunov2018} & 640 & 960 & \xmark & 0 & 43 & 2 & - & 4\\
        UADFV~\cite{Yang2019} & 49 & 98 & \xmark & 0 & 49 & 1 & - & 6\\
        FF++ DF~\cite{roessler2019faceforensicspp} & 4,000 & 5,000 & \xmark & 0 & ? & 4 & 2 & 19\\
        Google DFD~\cite{dufour_gully_2019} & 3,000 & 3,000 & \cmark & 28 & 28 & 5 & - & -\\
        Celeb-DF~\cite{Li2019} & 5,639 & 6,229 & \xmark & 0 & 59 & 1 & - & -\\
        DeeperForensics-1.0~\cite{Jiang2020} & 1,000 & 60,000 & \xmark & 100 & 100 & 1 & 7\tnote{c} & 5\\
        DFDC Preview\cite{Dolhansky2019} & 5,244 & 5,244 & \cmark & 66 & 66 & 2 & 3 & 3\\
        \hline
        \textbf{DFDC} & \textbf{104,500} & \textbf{128,154} & \cmark & \textbf{960} & \textbf{960} & \textbf{8}\tnote{d} & \textbf{19} &  \textbf{2,116}\\
        \hline
      \end{tabular}
      \begin{tablenotes}\footnotesize
        \item [a] The number of subjects who agreed to usage of their images and videos.
        \item [b] The number of publicly-available benchmark scores, from unique models or individuals. Due to the difficulty in finding all uses of a dataset, the scores must be in a centrally-located place (e.g. a paper or leaderboard). 
        \item [c] The DF-1.0 paper counts different perturbation parameters as unique. Our augmentations take real number ranges, making this number essentially infinite, so we only count unique augmentations, regardless of parameters.
        \item [d] Different methods can be combined with other methods; for simplicity our 8 methods are DF-128, DF-256, MM/NN, NTH, FSGAN, StyleGAN, refinement, and audio swaps.
    \end{tablenotes}
    \end{threeparttable}
  \label{tab:dataset_comparison}
\end{table*}

Due to the nature of the pairwise auto-encoder style model used to produce the majority of Deepfake videos, and due to limited availability of source footage, previous datasets contain few videos and fewer subjects. Specifically, every pairwise swap between two identities requires retraining a single model, which in modern hardware takes about one day on a single GPU. However, as noted in~\cite{roessler2019faceforensicspp}, the scale of a dataset in terms of raw training videos or frames is critical to a detector's performance.

Li et al.~\cite{Li2019} break down previous datasets into two broad categories - \textit{first generation} datasets such as DF-TIMIT~\cite{Korshunov2018}, UADFV~\cite{Yang2019}, and FaceForensics++ DF (FF++ DF)~\cite{roessler2019faceforensicspp}, and the \textit{second generation}, containing datasets such as the Google DeepFake Detection Dataset~\cite{dufour_gully_2019}, Celeb-DF~\cite{Li2019}, and the DFDC Preview Dataset~\cite{Dolhansky2019}. In general, each generation improves over the previous one by increasing the number of frames or videos by an order of magnitude. However, the datasets in the first two generations all suffer from a small number of swapped identities, which can contribute to overfitting on those particular identities. In all cases, apart from (possibly) FF++ DF, there are fewer than 100 unique identities. The FF++ DF dataset indexes by video sequence rather than ID, so it is unclear how many unique IDs appear in the dataset.

Finally, we propose a \textit{third generation} of datasets that not only have more than an order of magnitude larger number of frames and videos than the second generation, and with better quality, but also with agreement from individuals appearing in the dataset. This generation would include the DFDC, as well as the recent DeeperForensics-1.0 (DF-1.0) dataset~\cite{Jiang2020}. We believe that future face-swapped datasets should seek agreement from individual participants in order to be useful to and ethical for the research community.

\textbf{First generation}: Datasets in this generation usually contain less than 1,000 videos and less than 1 million frames. In addition, these datasets generally do not represent that they have the rights to the underlying content or agreement from individuals in the dataset. UADFV, DF-TIMIT, FaceForensics++ all contain videos sourced from YouTube and perform face swaps between public individuals. Additionally, due to the small scale, models trained on datasets such as FaceForensics++ usually do not generalize to real Deepfake videos~\cite{mama_shi_2019}. 
 
\textbf{Second generation}: These datasets generally contain between 1 and 10 thousand videos and 1 and 10 million frames, and contain videos with better perceptual quality than videos in the first generation. The Celeb-DF dataset contains over an order of magnitude more data than previous datasets, but contains data with use restrictions. During this generation, ethical concerns of subjects appearing in a dataset without their consent were publicly raised~\cite{solon_2019}. As a response, the preview version of this dataset, with consenting actors, was released. Shortly thereafter, the similar Google-DFD Dataset was also released, and also contained 28 paid actors. However, the datasets in this generation also do not contain enough identities that allow for sufficient detection generalization.
 
\textbf{Third generation}: The most recent DeepFake datasets, DeeperForensics-1.0 and the DFDC Dataset, contain tens of thousands of videos and tens of millions of frames. The encouraging trend of using paid actors has continued for both of these datasets. However, there are many major differences between DF-1.0 and the DFDC Dataset, and as DF-1.0 is the most similar dataset to the DFDC dataset, they are covered in detail.


First, although it is claimed that all videos in DF-1.0 contain consenting individuals, the target videos in the released dataset are all sourced from the internet, and whether these videos can be used freely is unclear. In addition, we do not count a fake video and the same fake video with a perturbation as two separate fake videos, as is done in DF-1.0, as these perturbations are trivial to add and do not require large scale computing resources. Every one of the 100,000 fake videos in the DFDC Dataset is a unique target/source swap. Ignoring perturbations, DF-1.0 only contains 1,000 unique fake videos. 

Furthermore, there are several notable differences besides raw numbers that have implications for the generalization performance of models trained on these datasets. The first is in the nature of the source data. DeeperForensics contains videos recorded in a controlled, studio setting, while the DFDC Dataset contains videos of individuals in indoor and outdoor settings, in a variety of real-world lighting conditions. The methods used to generate our dataset are flexible enough to handle this variety, and do not require frontally-captured videos taken in a studio. In addition, the DF-1.0 training dataset only contains videos produced by a single model that is proposed by the authors (and thus has not been used to create any public Deepfake videos), limiting the applicability of training on this set. 

Finally, in order to avoid the bias introduced by knowing how our manipulated data was produced, we do not propose any detection model trained specifically on our data, and instead solicited the community to contribute models that run on a hidden test set. Therefore, we constructed both a public test set (containing 4,000 videos) and a private test set (containing 10,000 videos), and included real "in-the-wild" Deepfakes. DF-1.0 uses a hidden test set of 400 videos, but it is not clear how many are real or fake or even whether or not they are Deepfaked videos. Finally, the perturbations used in DF-1.0 to expand the original set of 1,000 fake videos only contain basic pixel-level distortions such as color changes and Gaussian noise, and no semantic distractors that are present in real videos.

\section{DFDC Dataset}
\subsection{Source data}
Many Deepfake or face swap datasets consist of footage taken in non-natural settings, such as news or briefing rooms. More worryingly, the subjects in these videos may not agreed to have their faces manipulated.

With this understanding, we did not construct our dataset from publicly-available videos. Instead, we commissioned a set of videos to be taken of individuals who agreed to be filmed, to appear in a machine learning dataset, and to have their face images manipulated by machine learning models. In order to reflect the potential harm of Deepfaked videos designed to harm a single, possibly non-public person, videos were shot in a variety of natural settings without professional lighting or makeup, (but with high-resolution cameras, as resolution can be easily downgraded). The source data consisted of:

\begin{enumerate}
    \item 3,426 subjects in total with an average of 14.4 videos each, with most videos shot in 1080p
    \item 48,190 total videos that average 68.8s each - a total of 38.4 days of footage
    \item Over 25 TB of raw data
\end{enumerate}

The source videos were pre-processed with an internal face tracking and alignment algorithm, and all face frames were cropped, aligned, and resized to 256x256 pixels. For Deepfake methods, a subsample of 5,000 face frames collected from all videos was used to train models.

\subsection{Methods}
Throughout this section, the terms \textit{target} and \textit{source} are used. In general, \textit{target} refers to the base video in which a face will be swapped; \textit{source} refers to the source content that is used to extract the identity that will be swapped onto the target video. For example, for face swapping, we wish to put the \textit{source} face onto the \textit{target} face, resulting in a video identical to the \textit{target} video, but with the \textit{source} identity.

All of the face-swapped videos in the dataset were created with one of N methods. The set of models selected were designed to cover some of the most popular face swapping methods at the time the dataset was created. In addition, some methods with less-realistic results were included in order to represent low-effort Deepfakes. The number of videos per-method are not equal; the majority of face-swapped videos were created with the Deepfake Autoencoder (DFAE). This choice was made as to reflect the distribution of public Deepfaked videos, which are usually created with off-the-shelf software like DeepFaceLab\footnote{https://github.com/iperov/DeepFaceLab} or other public repositories used for creating Deepfakes. For a full description of each model's architecture, please refer to the Appendix.

\textbf{DFAE:} The DFAE model does not have a consistent name or design in public versions, but it is generally structured like a typical convolutional autoencoder model with several small but important differences. First, the model uses one shared encoder, but two separately-trained decoders, one for each identity in the swap. Additionally, the shared portion of the encoder extends one layer beyond the bottleneck, and the upscaling functions used are typically PixelShuffle operators, which is a non-standard, non-learned function that maps channels in one layer to spatial dimensions in the next layer. This architecture encourages the encoder to learn common features across both identities (like lighting and pose), while each decoder learns identity-specific features. At inference time, a given input identity in a frame is run through the opposite decoder, thus producing a realistic swap. The model is flexible; in the DFDC dataset, we included models that used an input/output resolution of 128x128 and 256x256. All of the images in the banner are fake faces, produced by a DFAE at 128x128 input/output resolution.

\textbf{MM/NN face swap:} The next method performed swaps with a custom frame-based morphable-mask model. Facial landmarks in the target image and source image are computed, and the pixels from the source image are morphed to match the landmarks in the target image using the method described in \cite{huang2012}. The eyes and the mouth are copied from the original videos using blending techniques, and spherical harmonics are used to transfer the illumination. This method works best when both the target and source face expressions are similar, so we used a nearest-neighbors approach on the frame landmarks in order to find the best source/target face pair. In a video, this approach is immediately evident, but on a frame-per-frame basis, the results look more realistic and could fool detectors that only operate on individual frames.

\newpage

We included three additional models based on methods that incorporate GANs - the Neural Talking Heads (NTH) model, FSGAN, and a method utilizing StyleGAN.

\textbf{NTH:} The NTH~\cite{zakharov2019few} model is able to generate realistic talking heads of people in few- and one-shot learning settings. It consists of two distinct training stages: a meta-learning stage and a fine-tuning stage. In the meta-learning stage, the model learns meta-parameters by transforming landmark positions into realistic-looking talking heads with a handful of training images of that person. In the fine tuning stage, both the generator and the discriminator are initialized with meta-parameters and quick coverage to the state that generate realistic and personalized images after seeing a couple of images of a new person. 


A pre-trained model is fine-tuned with pairs of videos in the raw DFDC set: the land marking positions are extracted from the driving video and fed into the generator to produce images with the appearance of the person in the other video. 

\textbf{FSGAN:} The FSGAN method (fully described in \cite{nirkin2019}) uses GANs to perform face swapping (and reenactment) of a source identity onto a target video, accounting for pose and expression variations. FSGAN applies an adversarial loss to generators for reenactment and inpainting, and trains additional generators for face segmentation and Poisson blending.  For the DFDC, we generated FSGAN swap videos using generator checkpoints trained on the data described in \cite{nirkin2019}, after performing brief fine-tuning on the reenactment generator for each source identity.

\textbf{StyleGAN:} The StyleGAN \cite{Karras2018ASG} method is modified to produce a face swap between a given fixed identity descriptor onto a video by projecting this descriptor on the latent face space. This process is executed for every frame.

\begin{figure}
  \centering
  \begin{subfigure}[b]{0.24\linewidth}
    \includegraphics[width=\linewidth]{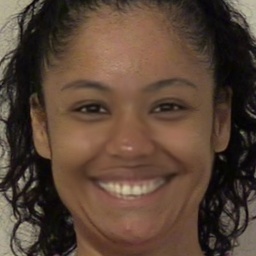}
  \end{subfigure}
  \begin{subfigure}[b]{0.24\linewidth}
    \includegraphics[width=\linewidth]{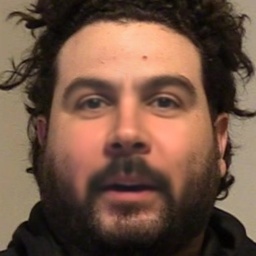}
  \end{subfigure}
  \begin{subfigure}[b]{0.24\linewidth}
    \includegraphics[width=\linewidth]{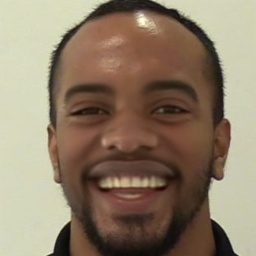}
  \end{subfigure}
  \begin{subfigure}[b]{0.24\linewidth}
    \includegraphics[width=\linewidth]{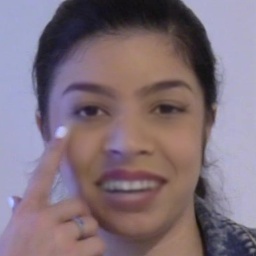}
  \end{subfigure}\\
  \begin{subfigure}[b]{0.24\linewidth}
    \includegraphics[width=\linewidth]{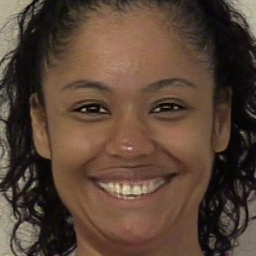}
  \end{subfigure}
  \begin{subfigure}[b]{0.24\linewidth}
    \includegraphics[width=\linewidth]{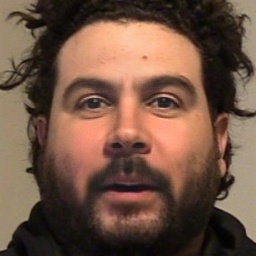}
  \end{subfigure}
  \begin{subfigure}[b]{0.24\linewidth}
    \includegraphics[width=\linewidth]{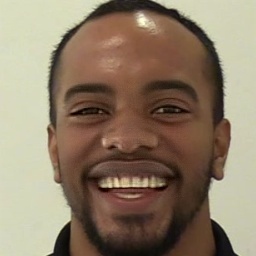}
  \end{subfigure}
  \begin{subfigure}[b]{0.24\linewidth}
    \includegraphics[width=\linewidth]{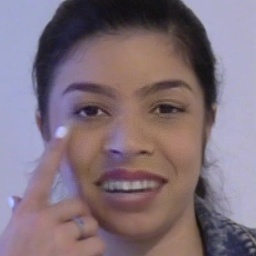}
  \end{subfigure}
  \caption{The fake faces in the top row were post-processed with a sharpening filter, resulting in the images in the bottom row.}
  \label{fig:refinement}
\end{figure}

\textbf{Refinement:} Finally, a random selection of videos went through post processing. Using a simple sharpening filter on the blended faces greatly increased the perceptual quality in the final video with nearly no additional cost, as shown in Figure~\ref{fig:refinement}. 



\subsubsection{Audio}

In addition to these face-swapping methods, we also performed audio swapping on some video clips using the TTS Skins voice conversion method detailed in \cite{polyak2019tts}. Video clips were selected for audio swapping independent of whether they were Deepfakes, or which face-swapping method was used. The voice identity used to swap did not depend on the face identity used in a swap. Although these audio manipulations were not considered 'Deepfakes' for the competition, we included them in the set to provide more breadth of manipulation type, and they may be of use in further research.

\subsection{Training}
All subjects were split into one of four sets: training, validation, public test, or private test. The training and validation sets were released publicly, while the public and private test sets were not released, as they were used to rank the final scores of all submissions to the Kaggle contest.

Not all possible pairs were trained. For example, an initial set of 857 subjects were selected into the training set, and there are over 360,000 potential pairings within this group. Training all pairs would require almost 1,000 GPU-years, assuming that it takes one day to train a DFAE model on one GPU. Instead, within a set, subjects were paired with those with similar appearances, as this tended to give better results for models like the DFAE. Over 800 GPUs were used to train 6,683 pairwise models (which required 18 GPU-years), as well the more flexible models such as NTH or FSGAN that only required a small amount of fine-tuning per subject. Finally, a subset of 10 second clips were selected from the output of all models, and the overall distribution of gender and appearance was balanced across all sets and videos.

\subsection{Post processing}

After inference, all methods produced a cropped image containing the face at 256x256 resolution. However, some methods do not infer details around the face, such as hair or background information. Therefore, we re-blended the face onto the original full-resolution raw frame using several steps, and combined the original raw frames and audio with \texttt{ffmpeg}.

First, we created a face mask using detected landmarks. The mask produced by these landmarks included the forehead region - many off-the-shelf algorithms only use a mask that extends to the eyebrow region, but this can lead to blending artifacts where "double eyebrows" appear in the final video. Next, we blended the face using the mask onto the original frame using Poisson blending. However, we did not use Poisson blending over the entire mask, as this would often blend the two identities and create an "average" face rather than a face that looks like the source subject. Instead, we only blended a small region along the edges of the mask. Practically, this was done using a set of morphological operations that extracted the mask border, applying a Gaussian filter to the mask border pixels, and finally Poisson blending the original and swapped face frames using this transformed mask.

Finally, it is important to note that proper face alignment enhanced the quality of all methods. Having each face aligned at a similar scale allowed models to focus on the details of a face, rather than having to rotate or translate mis-aligned faces. In addition, consistent alignment reduced face jitter in the final videos, which is usually a telltale sign that a video was Deepfaked. Faces were aligned by using a triangular set of positions formed by the two eyes and the nose, and computing an affine transform that best aligned a given face with these positions.

\begin{figure*}
\includegraphics[width=\linewidth]{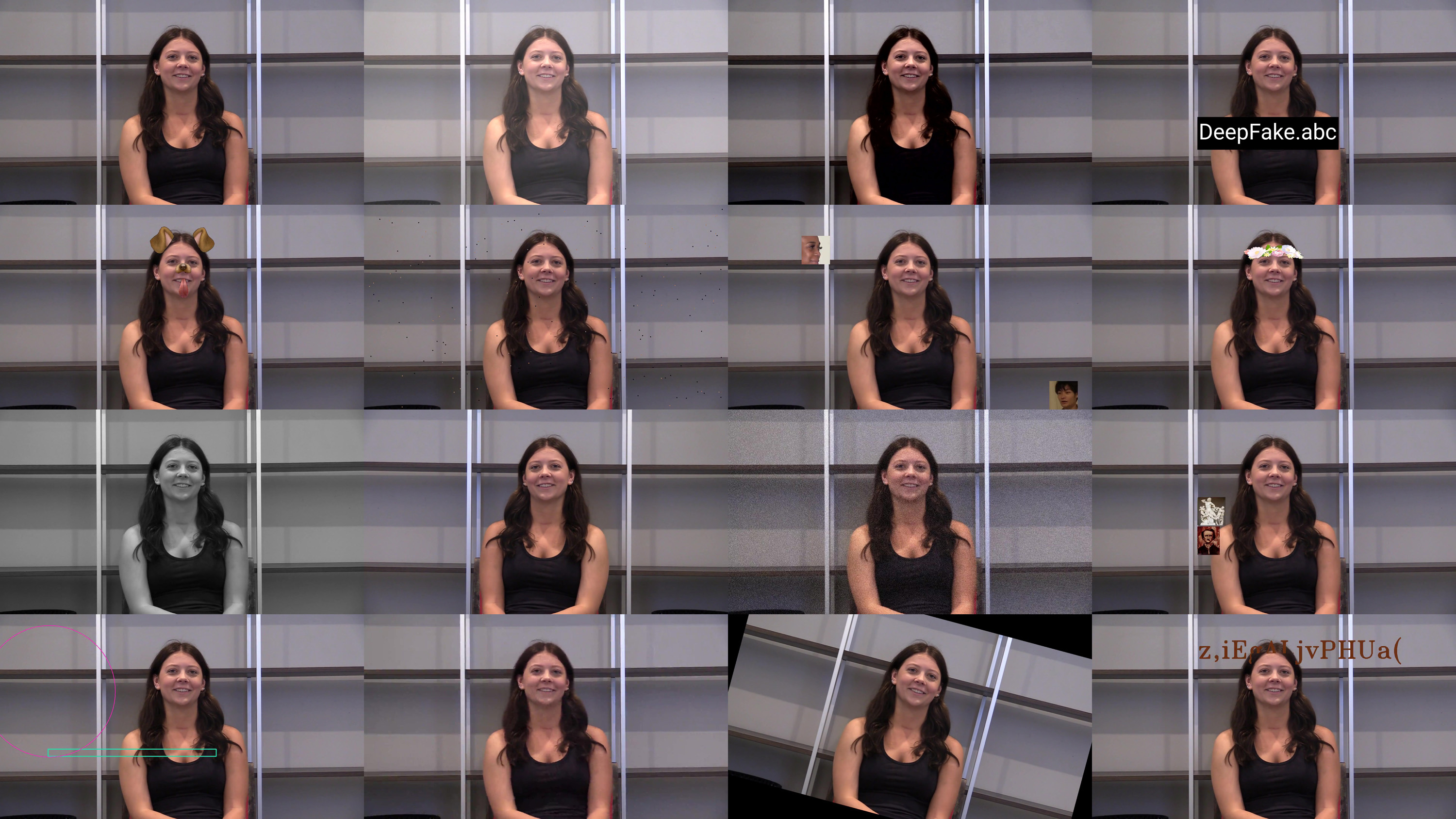}
\caption{A sample frame containing the various augmentations that were applied to the videos. Top row (from left to right): original, brightness, contrast, a logo overlay. Second row: dog filter, dots overlay, faces overlay, and flower crown filter. Third row: grayscale, horizontal flip, noise, and images overlay. Bottom row: shapes overlay, encoding quality level change, rotation, and text overlay. Not pictured: blur, framerate change, audio removal, and resolution change.}
\label{fig:augmentations}
\end{figure*}

\subsection{Dataset contents}

\textbf{Training set:} The training set provided was comprised of 119,154 ten second video clips containing 486 unique subjects. Of the total amount of videos, 100,000 clips contained Deepfakes which translates to approximately 83.9\% of the dataset being synthetic videos. In order to create the Deepfakes, the DFAE, MM/NN face swap, NTH, and FSGAN methods were used. No augmentations were applied to these videos.

\textbf{Validation:} The validation set is the public test set used to compute the public leaderboard positions in the Kaggle competition. This dataset consisted of 4,000 ten second video clips, in which 50\% (2000 clips) included Deepfakes. 214 unique subjects were used, none of which were a part of the training set. Additionally, the dataset included one unseen generation method for Deepfakes: StyleGAN. Augmentations were applied to roughly 79\% of all videos. 

\textbf{Test:} The private test set used was comprised of 10,000 ten second clips. Similar to the public test set, 50\% of the clips included Deepfakes and the other 50\% were non-Deepfaked clips. However, unlike the public test set, 50\% of this dataset includes organic content found on the internet and the other 50\% is unseen content from our source video dataset, collected from various sources. We are releasing the DFDC-like data portion from the private test set as the final test set.



The half of the final evaluation test set consisting of DFDC videos was assembled using 260 unique subjects from the source video dataset that have not been seen before. The data was constructed identically to our public test set, including all listed model types except for StyleGAN. Augmentations were applied to approximately 79\% of all videos in the final evaluation test set. New, never-before-seen augmentations were applied including a dog mask and a flower crown filter.

 Training, validation and testing corpuses can be downloaded from \url{http://ai.facebook.com} (URL to be updated).

\subsection{Augmentations}

Various augmentations such as geometric transforms or distractors were added to the videos in both the public Kaggle test set as well as the final evaluation test set. We defined two overarching types of augmentations:

\begin{enumerate}
    \item Distractor: overlays various kinds of objects (including images, shapes, and text) onto a video
    \item Augmenter: applies geometric and color transforms, frame rate changes, etc. onto a video
\end{enumerate}

Augmenters were randomly applied to approximately 70\% of the videos and are fairly straightforward transforms. The following types of augmenters were applied, all at randomly chosen levels: Gaussian blurring, brightening/darkening, adding contrast, altering the framerate, converting to grayscale, horizontal flipping, audio removal, adding noise, altering the encoding quality, altering the resolution, and rotating. All augmenters were present in both the public and final test sets, except for the grayscale augmenter which was only present in the final evaluation test set.

About 30\% of all videos contained distractors, some being more adversarial than others. The simplest distractors overlay random text, shapes, and dots onto each frame of a video and move around frame to frame. There can either be consistent movement (i.e. moving across horizontally or vertically) or random movement. The next subset of distractors overlay images onto each frame of a video. Similar to the previous subset, there can either be consistent or random movement. Additionally, there is an option to have the same image moving around the entire video, or the option to choose a random image every frame. Some of the added images included faces from the DFDC dataset. The last subset of distractors are facial filters that are commonly used on social media platforms. The facial filters implemented were the dog and flower crown filters.

All distractors were present in the final evaluation test set, however only the text, shapes, and faces distractors were present in the public Kaggle test set. The Deepfake YouTube channels logo distractor was only applied to benign videos in order to detect if any models were overfitting to YouTube data, which was not allowed by the competition's rules. See Figure \ref{fig:augmentations} for visual examples of the augmentations.
\newpage
\section{Metrics}\label{sec:metrics}

In most analyses of a machine learning model's performance, classification metrics such as log-loss are reported. In certain settings, this type of metric may be appropriate. For instance, a model trained to classify whether or not a video is a Deepfake may have its input pre-filtered by experts who want to use the model's score as an additional piece of evidence when performing a forensic analysis on a video. 

However, in many other settings, especially those in which an entity wants to find faked videos from a large set of input videos, \textit{detection} metrics are more appropriate. In this case, it is important to create metrics that reflect the true \textit{prevalence} (or the percentage of true positives) of a given type of fake video. In realistic distributions, the ratio of Deepfaked videos to real videos may be less than one in a million. With such an extreme class imbalance, accuracy is not as relevant as the precision or false positive rate of a model - the number of false positives of even an extremely accurate model will outnumber the true positives, thus decreasing the utility of the detection model.

\subsubsection*{Weighted PR}

\begin{figure*}
    \includegraphics[width=\linewidth]{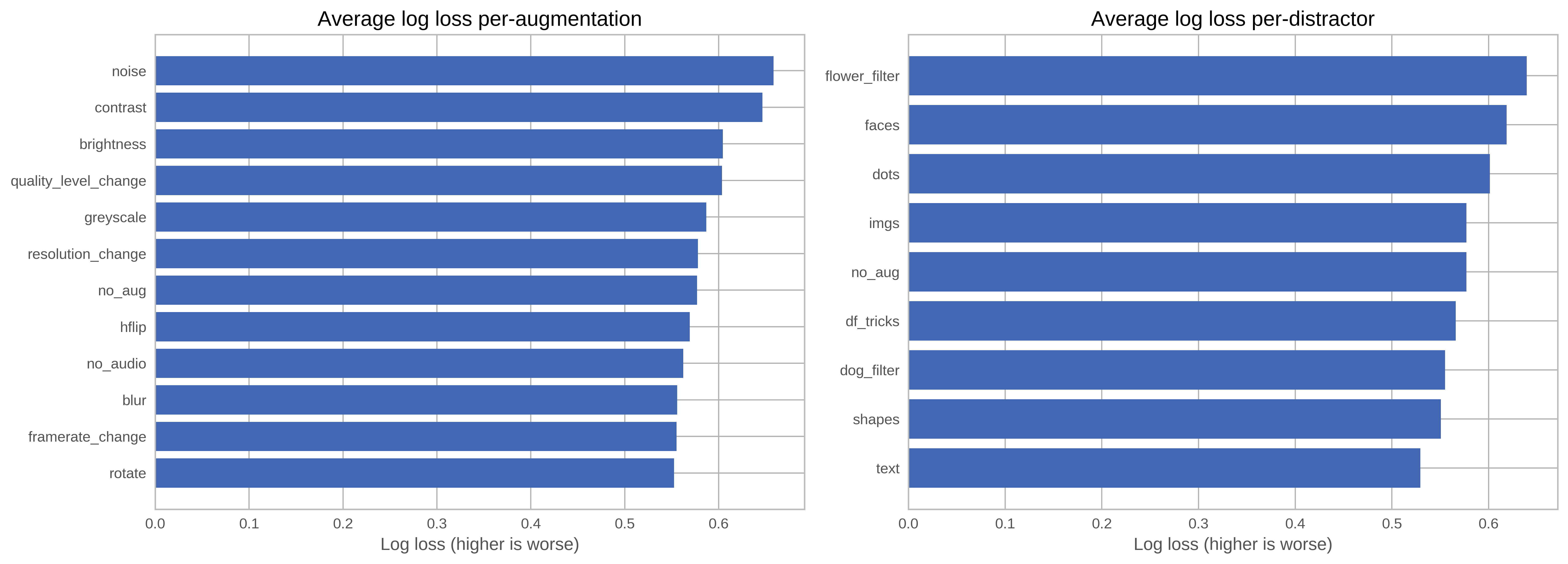}
    \caption{The average log loss for each augmenter and distractor. As expected, videos containing the noise and faces augmentations were among the most difficult to predict accurately. Surprisingly however, the flower filter, which only covers a portion of the forehead, was the most difficult distractor while the dog filter, which covers the nose and mouth, was one of the easier ones to predict. Horizontal flips, blurring, and rotation were among the easiest augmenters, likely due to the fact that they are common data augmentations. Note that videos that contained both augmenters and distractors were excluded from this analysis.}
    \label{fig:aug_results}
\end{figure*}

\par If we assume that the ratio between Deepfake and unaltered videos is $1:x$ in organic traffic and $1:y$ in a Deepfakes dataset, it is likely that $x\gg y$. Given the large number of true negatives, it is important for an automatic detection model to be precise. Even a model with high accuracy will detect many more false positives than true positives, simply because of the large class imbalance, which diminishes the utility of a automatic detection model. Metrics like the $F_\beta$ score (which does not weight false positives) or even the false positive rate (which only measures the tradeoff between true negatives and false positives) do not capture this issue. Instead, precision (or its inverse, the false discovery rate) is the most indicative metric for how a detection model will perform over a real distribution of videos. However, it is not practical to construct a dataset that mimics the statistics of organic traffic due to the sheer number of videos.

\begin{figure}
    \includegraphics[width=\linewidth]{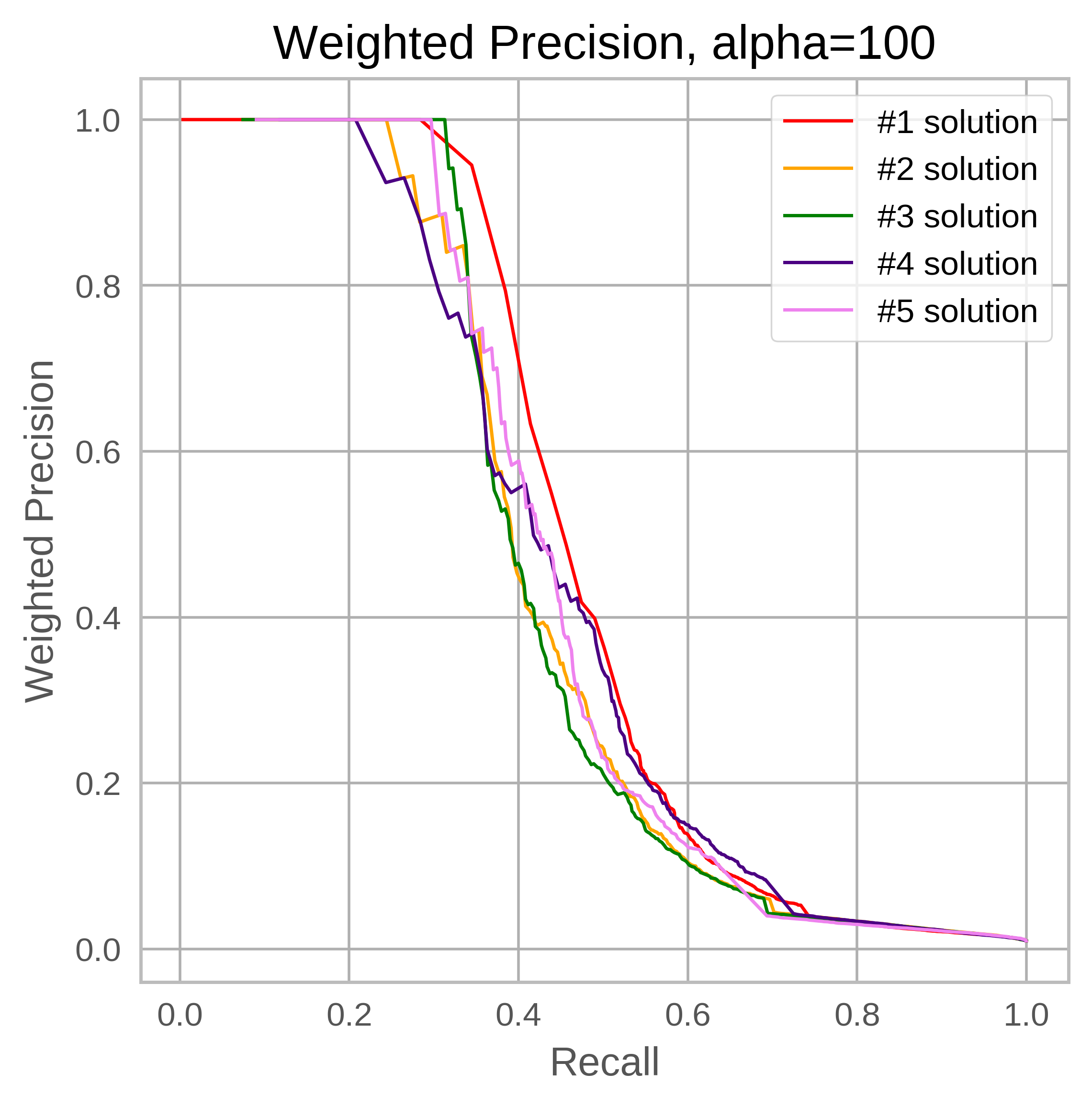}
    \caption{Weighted P/R curve, as described in Section~\ref{sec:metrics}. We set $\alpha=100$, which weights false positives 100x more than usual when calculating precision, and is designed to represent a more realistic distribution of DeepFake videos. Note that performance drops precipitously as more false positives are encountered.}
    \label{fig:weighted_pr}
\end{figure}

We can define a \emph{weighted precision} for a Deepfakes dataset as a very rough approximation of the precision that would be computed by evaluating on a dataset equal in size to the magnitude of organic traffic. Assuming the ratios of unaltered to tampered videos differ between a test dataset and organic traffic by a factor of $\alpha = x/y$, we define weighted precision wP and (standard) recall R as
\begin{equation}
\textrm{wP} = \frac{\textrm{TP}}{\textrm{TP}+\alpha\textrm{FP}},\;\;\;\;\;\; \textrm{R} = \frac{\textrm{TP}}{\textrm{TP}+\textrm{FN}}, 
\end{equation}
where TP, FP, and FN signify true positives, false positives, and false negatives. This metric differs from the $F_\beta$ score, as it assigns a weight to false positives instead of false negatives (and ignores true negatives). For an example applied to submissions in the DFDC, see Figure~\ref{fig:weighted_pr}.

For the purposes of ranking submissions in the DFDC Competition, models were ranked by log loss, as the weighted PR metric can be extremely small and somewhat noisy. In addition, we only needed a relative measure of performance to rank competitors, rather than an absolute measure. However, in Section~\ref{sec:benchmarking}, we report weighted PR metrics in addition to log loss to assess each model's performance as a detector on a realistic distribution of videos.

\section{Results}\label{sec:results}


\begin{figure*}[ht]
    \includegraphics[width=\linewidth]{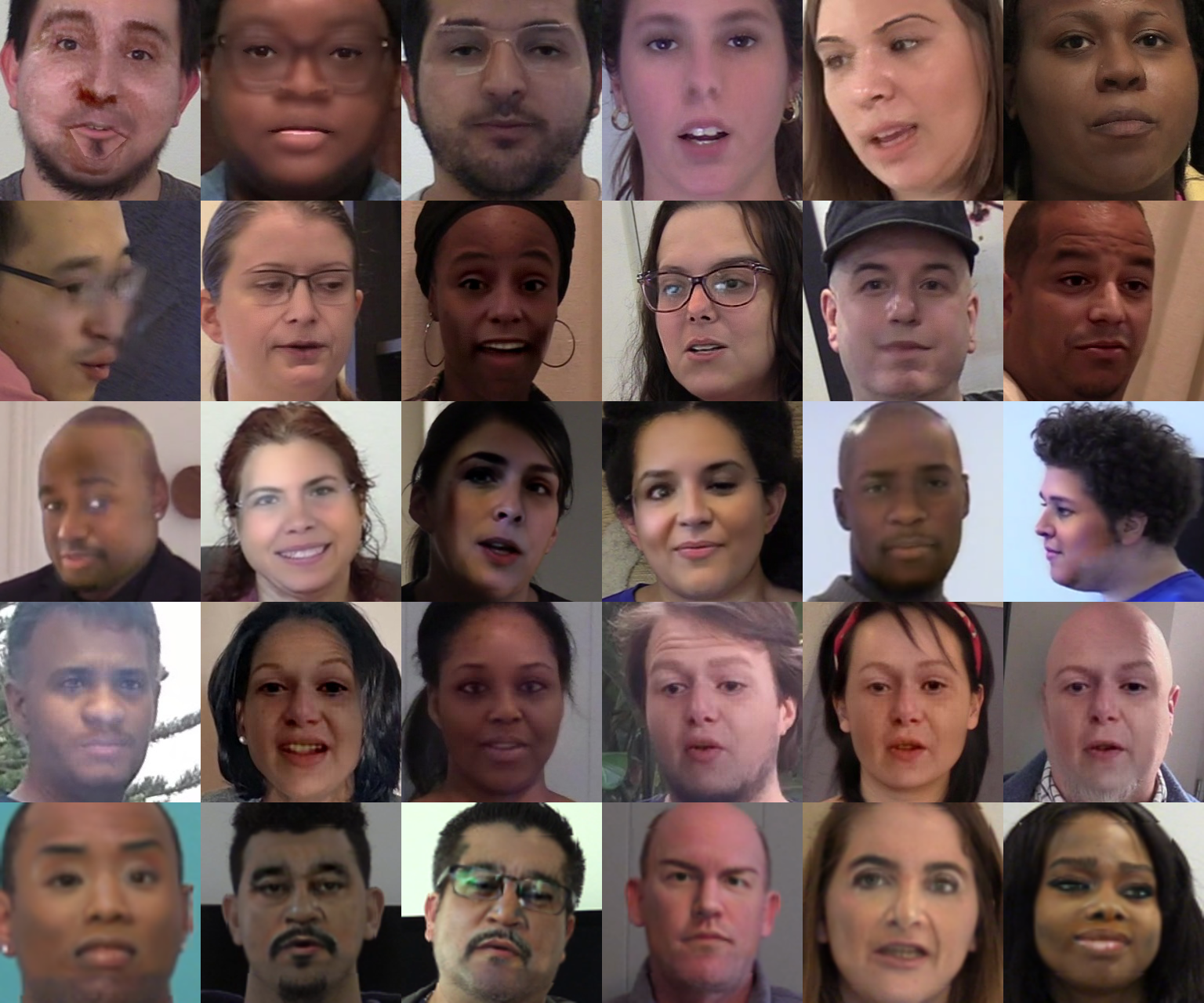}
    \caption{A selection of results of varying quality, where each row corresponds to a method (from top-to-bottom, MM/NN, DFAE, FSGAN, NTH, StyleGAN), and each column shows increasing quality from left to right. For a more detailed overview, see Section~\ref{sec:results}.}
    \label{fig:results}
\end{figure*}

As we do not introduce any novel architectures, in this section we describe how well different models and methods perform in practice, and show some of the best and worst examples of each method. Unlike other work in this area, we explicitly do \textit{not} show the worst examples from datasets other than the DFDC Dataset as a comparison, as (a) it is simple to cherry-pick the worst examples from a distribution of data produced by automatic methods, and (b) the perceptual quality of a moving video cannot be demonstrated with individual still frames. Instead, we believe that producing a very large dataset that covers a wide range of output qualities and with many unique videos is more useful than a hand-tuned, high-quality dataset of limited size.

In general, face swaps produced with DFAE methods were of higher quality over a wider range of videos than swaps produced with GAN-like methods, and required much less fine tuning. Our hypothesis is that GAN-like methods work well in limited settings with even lighting, such as news rooms, interviews, or controlled-capture videos as in \cite{Jiang2020}, but do not work well automatically (yet). Beyond ease of use, this may explain why most public Deepfake videos are produced with DFAE methods. Consequently, the majority of videos in the DFDC Dataset were produced with several different DFAE variants.

Some qualitative results from each method are shown in Figure~\ref{fig:results}, with further discussion here.

\textbf{MM/NN}: While this method was able to produce convincing single frame images, overall the NN approach tended to produce discontinuities in the face. In addition, there was sometimes a failure in the mask fitting, as seen in the left image of the top row of Figure~\ref{fig:results}.

\textbf{DFAE}: The DFAE methods were generally the most flexible and produced the best results out of the methods included in this paper. They were able to handle a variety of lighting conditions and individuals with good temporal coherence, even though inference happened on a frame-by-frame basis. Particular areas of weakness were glasses and extreme poses.

\textbf{FSGAN}: The FSGAN was able to produce convincing results in scenes with good lighting, but struggled to maintain an even skin tone in darker settings. In addition, it tended to produce flat-looking results. One particular strength of this method is in handling extreme head poses, as shown in the rightmost image of row 3 of Figure~\ref{fig:results}.

\textbf{NTH}: Of the GAN like methods, this method produced the most consistent quality. However, it tended to insert similar looking eyes across subjects, regardless of the source ID. Like other GAN methods, NTH did not produce good results in darker settings.

\textbf{StyleGAN}: Overall, StyleGAN produced the worst overall results, both at the frame level and at the video level. By far the most common in issue in videos was an unconstrained eye gaze. Without conditioning on the input gaze, the gaze of the swapped face tender to wander, with eyes looking in different directions at once. In addition, StyleGAN had trouble matching the illumination in a scene.
\section{Large scale benchmarking}\label{sec:benchmarking}
The second component of this work involved a large public competition, where participants submitted Deepfake detection models trained on the full DFDC Dataset. Initially, the public test set was used to rank the public leaderboard while the competition was ongoing. This set only contained DFDC videos with subjects that never appeared in the dataset. The "private" test set included real videos, some of which were Deepfakes, in addition to more DFDC videos that contained even more subjects that hadn't appeared in any previous set. Participants were free to use additional external data, as long as it complied to the policies of the competition.

The following analysis presents a comprehensive snapshot of the current performance of Deepfake detectors, and in particular, the performance against the private test set gives an idea as to how the best models would perform on a real video distribution.

\subsection{Meta analysis}
\begin{figure}[h!]
  \centering
    \includegraphics[width=\linewidth]{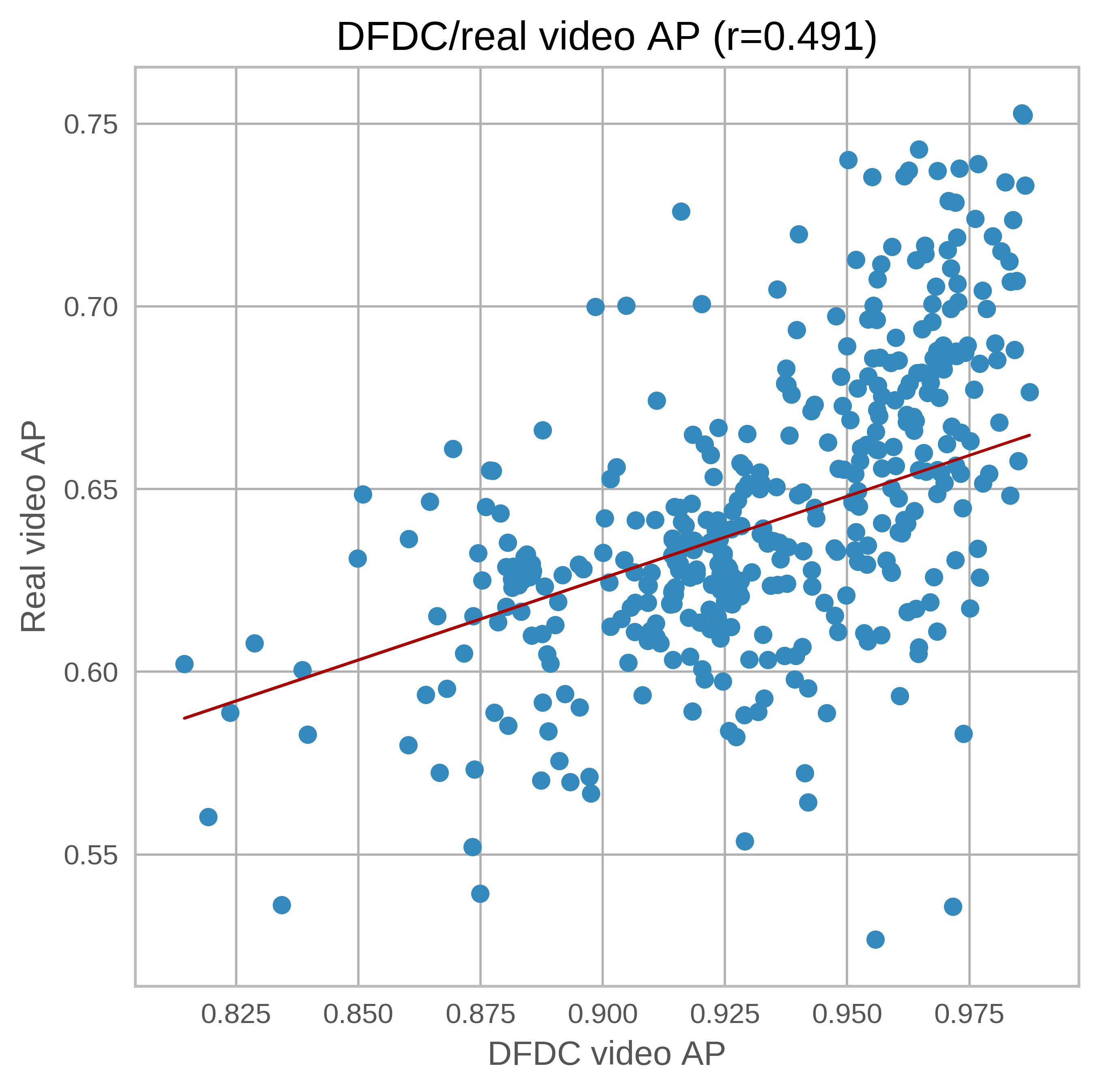}
  \caption{Correlation between the average precision of submitted models for DFDC and real videos.}
  \label{fig:corr}
\end{figure}

\begin{figure}
    \includegraphics[width=\linewidth]{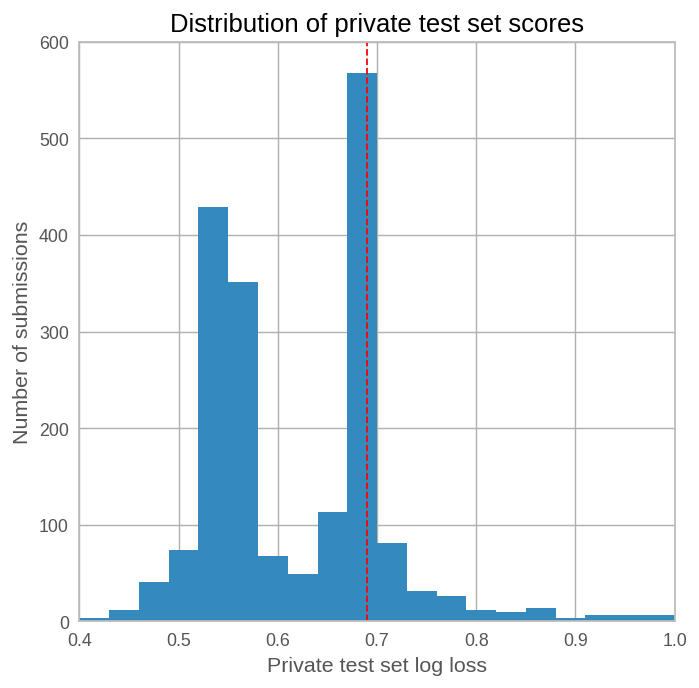}
    \caption{Distribution of private test set log loss scores. The vertical line indicates random performance (i.e. predicting 0.5 for every video).}
    \label{fig:pts_distr}
\end{figure}

During the course of the competition, 2,114 teams participated.  Teams were allowed to submit two different submissions for final evaluation. Of all of the scores on the private test set, 60\% of submissions had a log loss lower than or equal to 0.69, which is roughly the score if one were to predict a probability of 0.5 for every video. As seen in Figure~\ref{fig:pts_distr}, many submissions were simply random. Good performance on the public test set correlated with good performance on the private test set, as shown in the first image of Figure~\ref{fig:corr}. 

All final evaluations were performed on the private test set, using a single V100 GPU. Submissions had to run over all 10,000 videos in the private test set within 90 hours, but most submissions finished evaluating all videos within 10 total hours, giving a rough average inference time of around 3.6s per-video.

\subsection{Analysis of submitted models}
\begin{table*}[t]
  \centering
  \begin{threeparttable}
    \caption{Top 5 model results - precision reported at recall levels 0.1, 0.5, and 0.9}
      \begin{tabular}{|l|rrrrrr|}
      \hline
      Team name & \shortstack[r]{Overall\\log loss} & \shortstack[r]{DFDC\\log loss} & \shortstack[r]{Real\\log loss} & \shortstack[r]{Real\\P@0.1} & \shortstack[r]{Real\\P@0.3} & \shortstack[r]{Real\\P@0.9}\\
      \hline
      	Selim Seferbekov~\cite{Seferbekov2020}& 0.4279 & 0.1983 & 0.6605 & 0.9803 & 0.7610 & 0.5389\\
        WM~\cite{wm2020} & 0.4284 & 0.1787 & 0.6805 & 0.9294 & 0.6717 & 0.5775\\
        NTechLab~\cite{NTechLab2020} & 0.4345 & 0.1703 & 0.7039 & 0.9804 & 0.8244 & 0.5541\\
        Eighteen Years Old~\cite{Eighteen2020}& 0.4347 & 0.1882 & 0.6831 & 0.9843 & 0.6329 & 0.5625\\
        The Medics~\cite{Medics2020} & 0.4371 & 0.2157 & 0.6621 & 0.9653 & 0.7354 & 0.5516\\
      \hline
      \end{tabular}
    \end{threeparttable}
  \label{tab:winning_results}
\end{table*}

\begin{figure*}[ht]
  \centering
  \begin{subfigure}[b]{0.33\linewidth}
    \includegraphics[width=\linewidth]{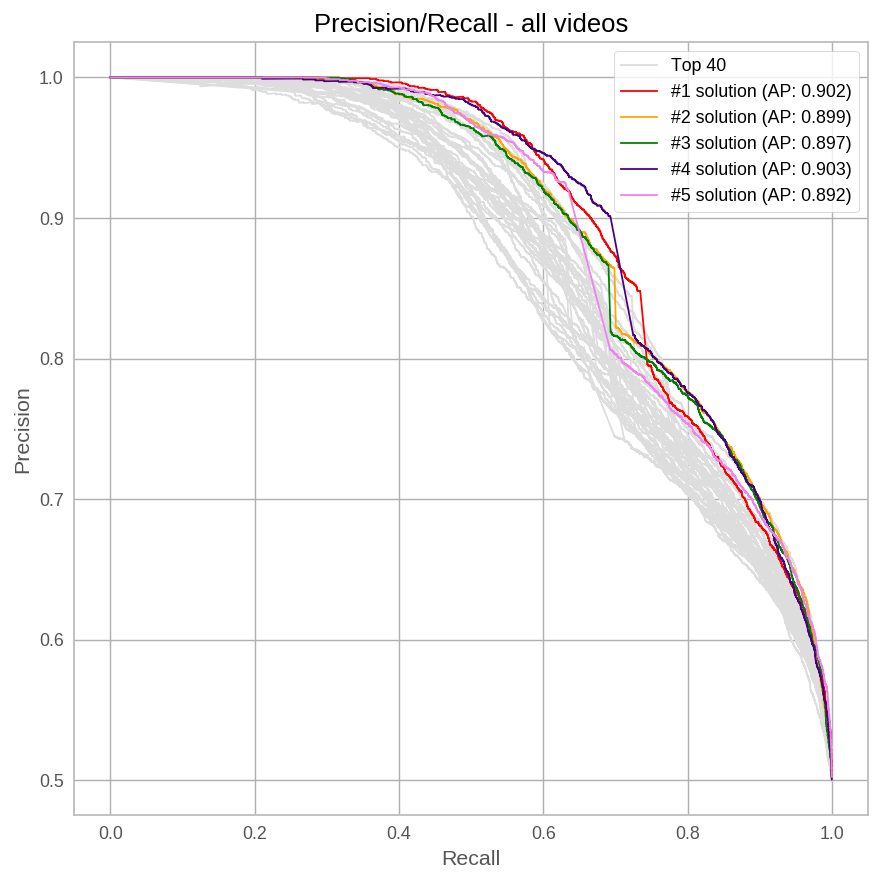}
  \end{subfigure}
  \begin{subfigure}[b]{0.33\linewidth}
    \includegraphics[width=\linewidth]{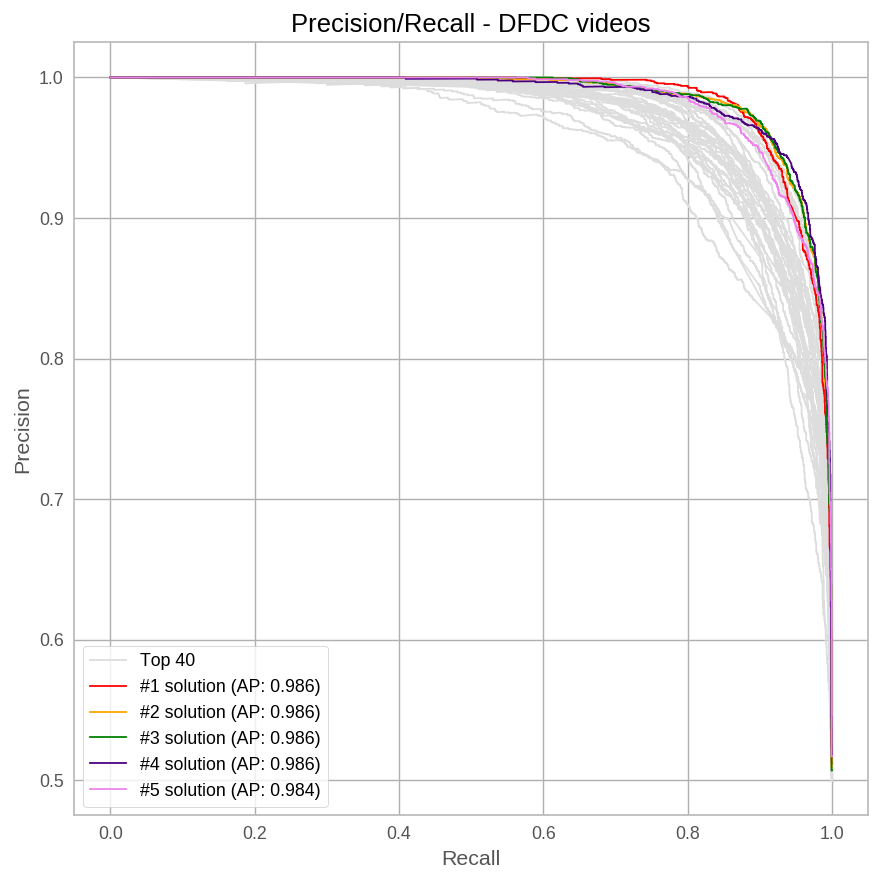}
  \end{subfigure}
  \begin{subfigure}[b]{0.33\linewidth}
    \includegraphics[width=\linewidth]{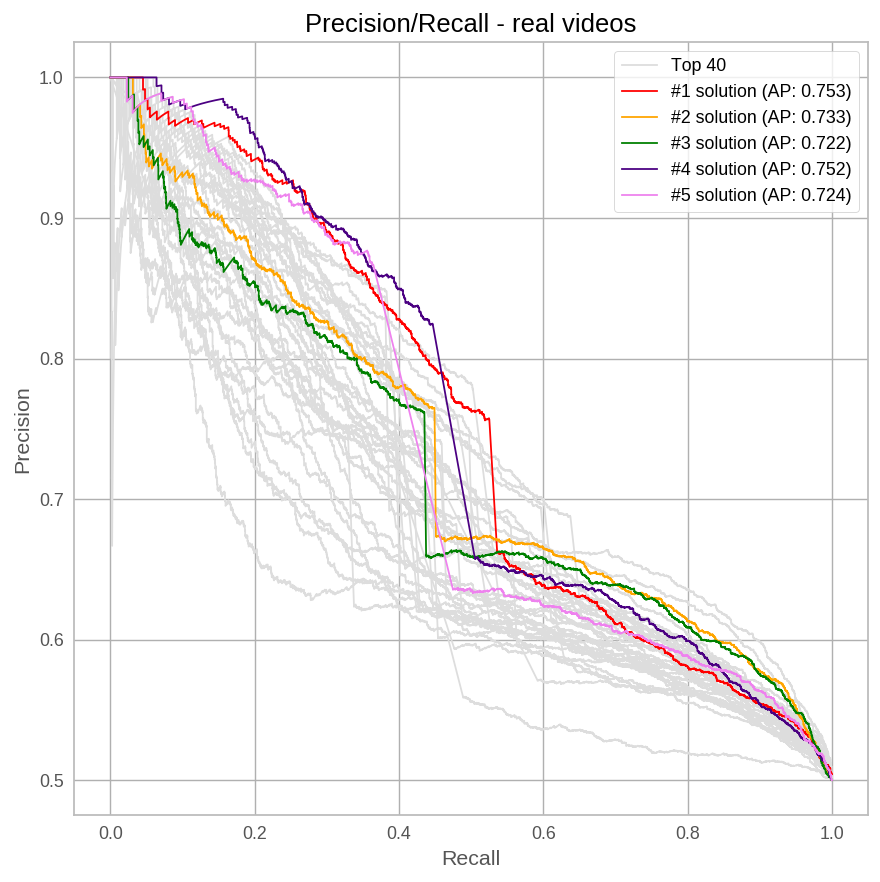}
  \end{subfigure}\\
  \begin{subfigure}[b]{0.33\linewidth}
    \includegraphics[width=\linewidth]{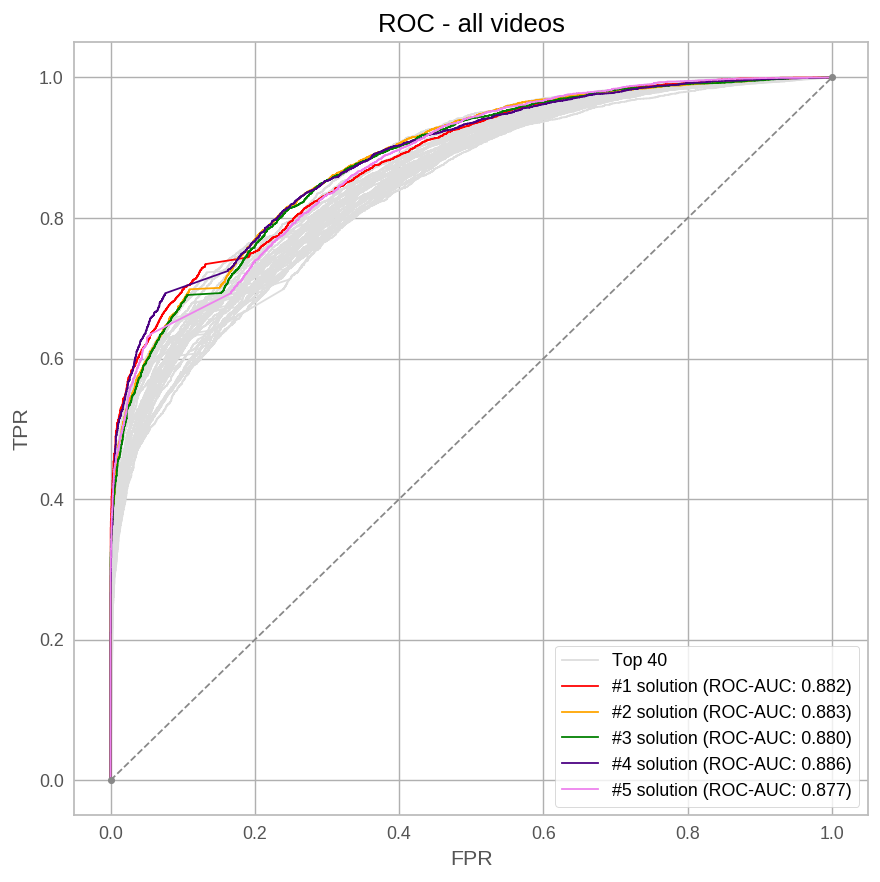}
  \end{subfigure}
  \begin{subfigure}[b]{0.33\linewidth}
    \includegraphics[width=\linewidth]{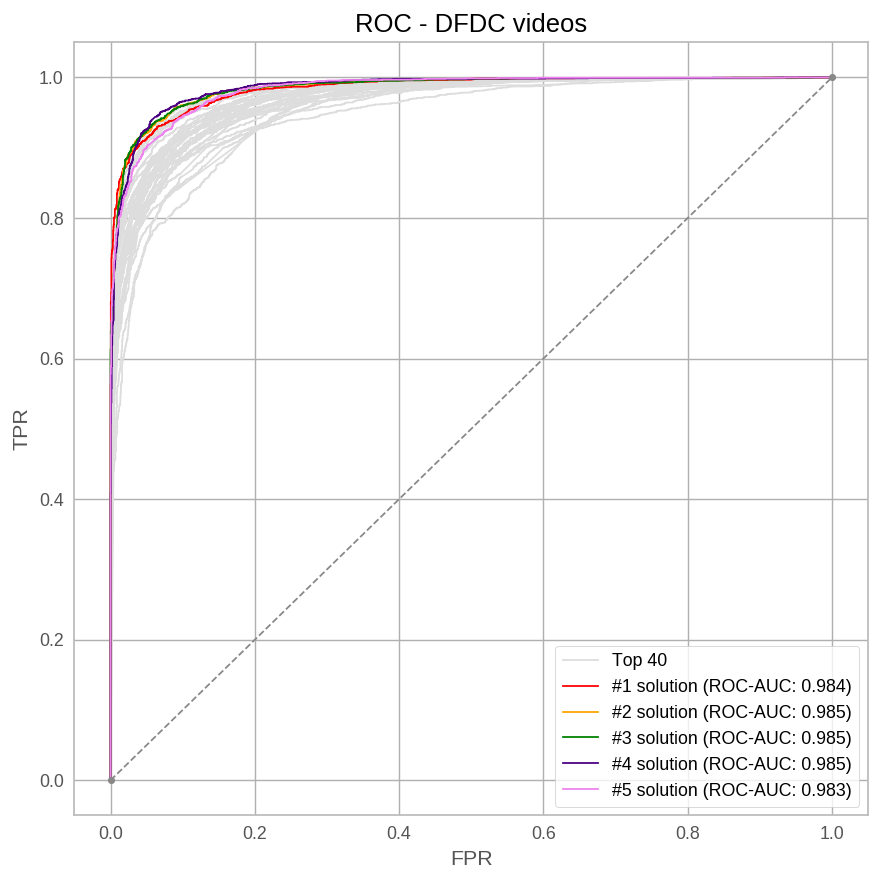}
  \end{subfigure}
  \begin{subfigure}[b]{0.33\linewidth}
    \includegraphics[width=\linewidth]{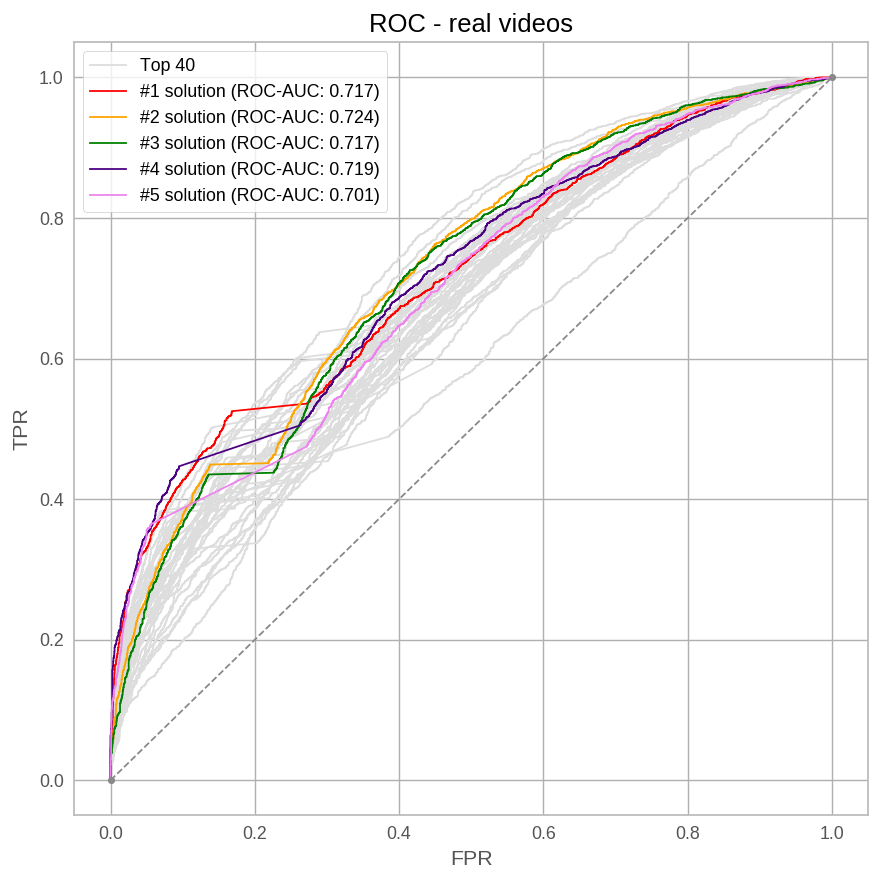}
  \end{subfigure}
  \caption{Detection metrics broken down by video type.}
  \label{fig:pr_roc}
\end{figure*}

After the competition ended, all scores for all submissions were computed over all videos in the private test set. Shown in Figure~\ref{fig:pr_roc} are detection metrics computed over the entire private set, only on videos from the DFDC, and only on real videos. As expected, the best models achieved very good detection performance on DFDC videos, as the all of the videos in the training set came from this distribution. On real videos, there was an expected performance drop, but the best models achieved an average precision of 0.753 and a ROC-AUC score of 0.734, only on real videos, which demonstrates that training on the DFDC Dataset allows a model to generalize to real videos. 

The second and third plots in figure~\ref{fig:corr} show the correlation between detection metrics computed on DFDC videos only and scores on real videos only, providing additional evidence that good performance on the DFDC dataset translates to good performance on real videos, and consequently that the DFDC dataset is a valuable resource for training real Deepfake detection models.

Finally, we provide a brief description of the top-5 winning solutions here - more detailed analysis of each approach can be found at the accompanying reference for each model. The first submission, \textit{Selim Seferbekov}~\cite{Seferbekov2020}, used MTCNN~\cite{zhang2016joint} for face detection and an EfficientNet B-7~\cite{Tan2019} for feature encoding. Structured parts of faces were dropped during training as a form of augmentation. The second solution, \textit{WM}~\cite{wm2020}, used the Xception~\cite{chollet2017xception} architecture for frame-by-frame feature extraction, and a WS-DAN~\cite{hu2019see} model for augmentation. The third submission, \textit{NTechLab}~\cite{NTechLab2020}, used an ensemble of EfficientNets in addition to using the \textit{mixup}~\cite{zhang2017mixup} augmentation during training. The fourth solution, \textit{Eighteen Years Old}~\cite{Eighteen2020}, used an ensemble of frame and video models, including EfficientNet, Xception, ResNet~\cite{he2016deep}, and a SlowFast~\cite{feichtenhofer2019slowfast} video-based network. In addition, they tailored a score fusion strategy specifically for the DFDC dataset. Finally, the fifth winning solution, \textit{The Medics}~\cite{Medics2020}, also used MTCNN for face detection, as well as an ensemble of 7 models, 3 of which were 3D CNNs (which performed better than temporal models), including the I3D model~\cite{carreira2017quo}.

\section{Future work}
There are three main areas of future work regarding the DFDC Dataset. First, we would like to perform a large scale perceptual study of the quality of the videos in the dataset. Due to time constraints and extenuating circumstances surrounding COVID-19, this portion of the project is delayed, but is ongoing. Second, we would like to expand the overall size of the dataset. Only 960 of the roughly 3,500 original identities were included in the dataset, again due to time and computational constraints. Finally, we are exploring the possibility of releasing the original raw dataset to the research community. One of the main differences with previous Deepfake datasets is that they do not purport to have agreement from individuals to be included in the datasets. Releasing all of the roughly 50k 1 minute videos with some additional annotations will help alleviate this problem, and hopefully lead to even higher quality and larger Deepfake datasets in the future.

{\small
\bibliographystyle{ieee_fullname}
\bibliography{egbib}
}

\end{document}